\documentclass[10pt,twocolumn,letterpaper]{article}

\usepackage[pagenumbers]{cvpr} 

\usepackage{graphicx}   
\usepackage{amsmath}    
\usepackage{amssymb}    
\usepackage{booktabs}   
\usepackage{tabularx}   
\usepackage{makecell}   
\usepackage{siunitx}    
\usepackage{caption}
\usepackage{threeparttable}
\usepackage{newtxtext, newtxmath}
\usepackage{amssymb}
\newcolumntype{C}{>{\centering\arraybackslash}X}
\newcolumntype{L}{>{\raggedright\arraybackslash}X}

\newcommand{\Sec}[1]{Sec.~\ref{#1}} 
\newcommand{\Fig}[1]{Fig.~\ref{#1}}
\newcommand{\Table}[1]{Table~\ref{#1}}

\newsavebox{\mybox}

\definecolor{cvprblue}{rgb}{0.21,0.49,0.74}
\usepackage[pagebackref,breaklinks,colorlinks,allcolors=cvprblue]{hyperref}

\def\paperID{3532} 
\def\confName{CVPR}
\def\confYear{2026}

\title{TDEdit: A Unified Diffusion Framework for Text-Drag Guided Image Manipulation}

\author{
    Qihang Wang\textsuperscript{1} \quad
    Yaxiong Wang\textsuperscript{1} \thanks{Corresponding author: wangyx15@stu.xjtu.edu.cn} \quad
    Lechao Cheng\textsuperscript{1} \quad
    Zhun Zhong\textsuperscript{1}\\
    \textsuperscript{1}Hefei University of Technology 
}

\begin{document}
\maketitle

\begin{figure*}[t]
    \centering
    \includegraphics[width=0.95\textwidth]{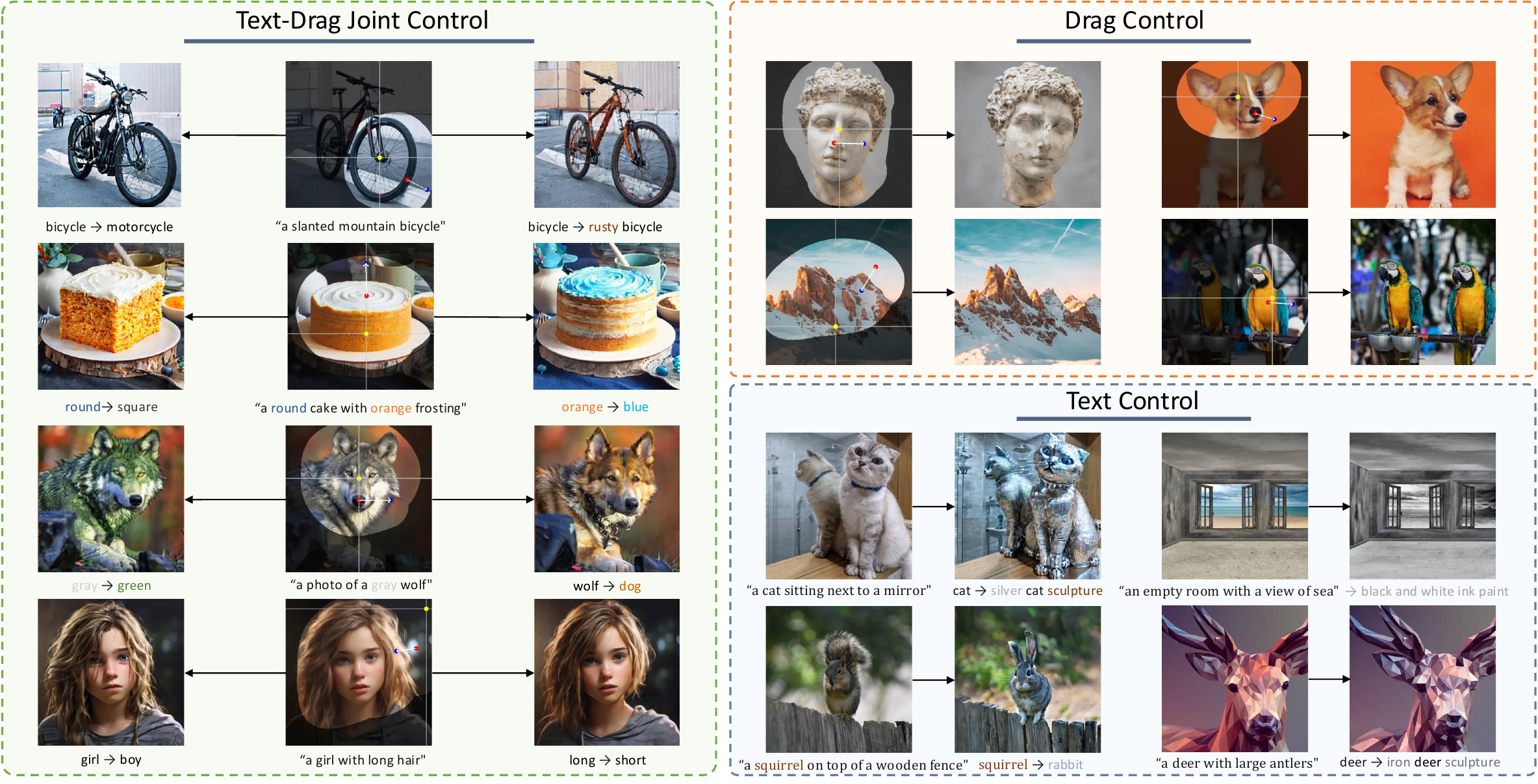}
    \caption{We present \emph{TDEdit}, a unified diffusion-based framework for text-drag guided image editing. TDEdit achieves versatile manipulations through three distinct control modalities: joint text-drag guidance (left), exclusive drag-based deformation (top right), and  pure textual control (bottom right).}
    \label{fig:teaser}
\end{figure*}

\begin{abstract}

  This paper explores image editing under the joint control of text and drag interactions. While recent advances in text-driven and drag-driven editing have achieved remarkable progress, they suffer from complementary limitations: text-driven methods excel in texture manipulation but lack precise spatial control, whereas drag-driven approaches primarily modify shape and structure without fine-grained texture guidance. To address these limitations, we propose a unified diffusion-based framework for joint drag-text image editing, integrating the strengths of both paradigms. Our framework introduces two key innovations: (1) Point-Cloud Deterministic Drag, which enhances latent-space layout control through 3D feature mapping, and (2) Drag-Text Guided Denoising, dynamically balancing the influence of drag and text conditions during denoising. Notably, our model supports flexible editing modes—operating with text-only, drag-only, or combined conditions—while maintaining strong performance in each setting. Extensive quantitative and qualitative experiments demonstrate that our method not only achieves high-fidelity joint editing but also matches or surpasses the performance of specialized text-only or drag-only approaches, establishing a versatile and generalizable solution for controllable image manipulation. Code will be made publicly available to reproduce all results presented in this work.
\end{abstract}
\section{Introduction}






The advent of diffusion models~\cite{ddpm,ddim,latentdiff} has marked a paradigm shift in generative image editing, offering unprecedented control over content creation and manipulation. Building upon this foundation, two distinct yet powerful editing paradigms have emerged: text-driven~\cite{infedit,text2live,cycled,sdedit,prompt2prompt} and drag-driven~\cite{fastdrag,dragdiff,dragnoise,gooddrag,dragan} approaches. Text-driven methods, empowered by large-scale vision-language models like CLIP~\cite{clip}, enable high-level semantic edits through natural language instructions - allowing users to modify object attributes, swap textures, or even alter scene semantics with remarkable fidelity. Concurrently, drag-driven techniques have gained traction for their intuitive spatial manipulation capabilities, where users can deform, reposition, or rotate objects through simple point-and-drag interactions. While both approaches have demonstrated impressive results in their respective domains, they represent fundamentally different axes of control: one operating in the semantic space and the other in the geometric space.  

A closer examination reveals that these two editing models exhibit complementary strengths and limitations: Text-driven approaches, while excelling at texture modification like color changing, object replacement, often struggle with precise spatial control due to the inherent ambiguity of language descriptions and the limitations of cross-attention mechanisms in diffusion models. For instance, while a prompt like "make the dog fluffier" can effectively modify texture, attempts to specify exact poses or orientations through text alone frequently lead to inconsistent or unpredictable results~\cite{sdedit,diffEdit,prompt2prompt}. Conversely, drag-driven methods provide pixel-level precision for geometric transformations but lack the semantic understanding needed for coherent texture synthesis or attribute modification~\cite{dragdiff,freedrag,dragan}. This limitation becomes particularly apparent when users attempt to combine geometric changes with texture edits - a common requirement in practical editing scenarios. Nevertheless, current models only pay attention to either text-driven control or drag-based edit, which cannot meet the complex requirements in practical scenarios.


To address this limitation, we propose a unified editing framework capable of simultaneously processing drag-based and text-based controls while preserving the individual capabilities of each modality. Developing such a framework presents two key challenges. (1) Dynamic Guidance Balancing: The denoising process requires careful modulation between text and drag guidance to prevent one modality from dominating the other. Naively enforcing both conditions may lead to degraded control fidelity in either domain. (2) Geometric Consistency in Diffusion: Maintaining precise spatial alignment under hybrid conditions remains nontrivial, as the interplay between text and drag signals often disrupts fine-grained layout control—a critical issue that demands resolution.


To address dynamic guidance balancing, we propose the Drag-Text Guided Denoising (DTGD) mechanism. DTGD adjusts conditional clues based on denoising steps. To balance the control signals, DTGD uses a three-branch architecture~\cite{infedit}: a source branch provides the original layout and details; a reference branch maintains global layout and injects target prompt semantics; and a target branch preserves the dragged region's local layout while extracting details from the reference branch. During denoising, the target branch fuses semantic details from both branches using a dynamic factor for blended control.

To achieve precise layout control, we propose Point-Cloud Deterministic Drag (PCDD), a strategy for understanding drag intent using 3D-aware position representation. PCDD first projects 2D drag inputs into a 3D point cloud via depth-aware feature mapping. It then applies deterministic transformations to estimate post-drag pixel positions. The 2D coordinates are projected into 3D space, where rigid and non-rigid transformations are performed based on drag cues. Finally, the 3D coordinates are mapped back to 2D, adjusting latent features to align with the drag intent.

With the above designs as the main force, we finally develop our TDEdit: A Unified Diffusion Framework for Joint Text-Drag
Guided Image Manipulation.  In summary, we highlight the contributions of this paper as follows.

\begin{itemize}
    \item We make an early exploration for a text-drag joint control framework for image editing, and present an unified framework TDEdit that is capable of  simultaneously processing drag-based and text-based controls as well as the individual control capabilities of each modality.

    \item A Point-Cloud Deterministic Drag (PCDD) mechanism is proposed. PCDD distorts the latent feature via a 3D mapping to ensure the final layout of final image can precisely follow both conditions.

    \item To dynamically balance the control of text and drag instructions, we propose a Drag-Text Guided Denoising (DTGD) mechanism, offering a flexible trade-off between the text and drag conditions during denoising.
    
    \end{itemize}

\section{Related Work}

\begin{figure*}[t] 
\centering 
\includegraphics[width=\textwidth]{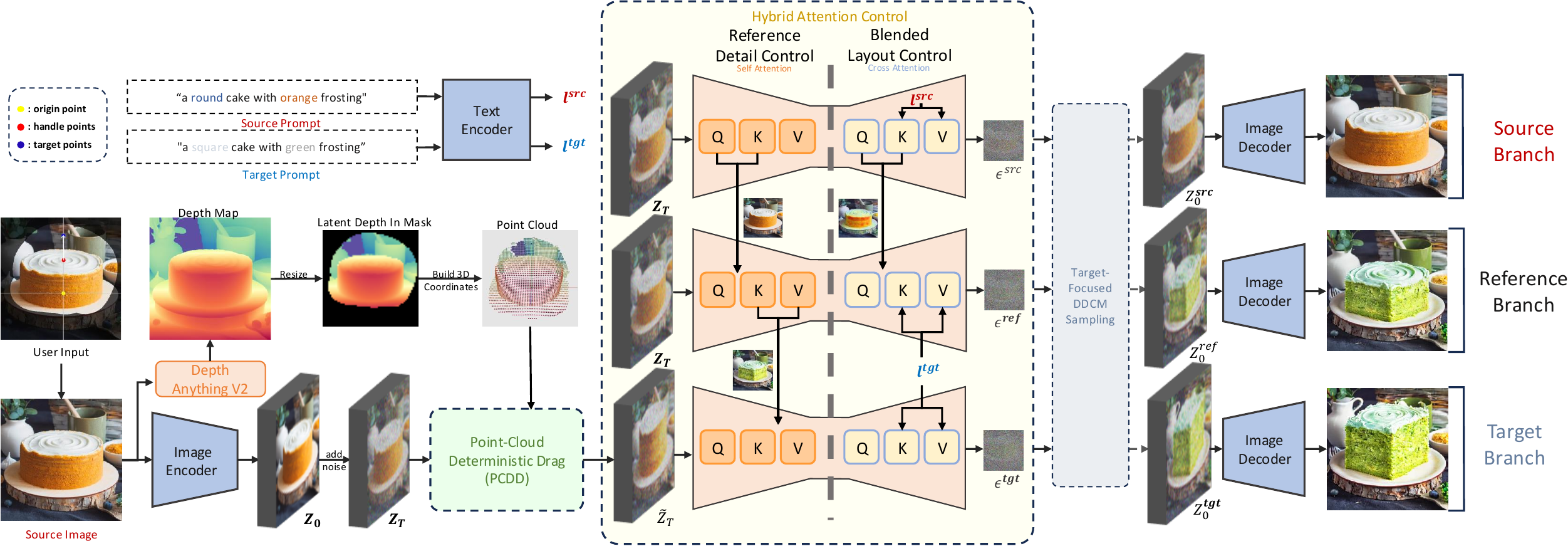} 
\caption{Illustration of our proposed TDEdit framework. The input image is first encoded via image encoder into latent feature. Next, PCDD utilizes the drag condition and estimated depth map to distort the latent features after noise addition. In the following, the distorted image latent and the original one as well as the prompts pass through DTGD  with three branches to acquire the final latent feature, which is then fed into the image decoder to harvest the edited image.} 
\label{TDEdit}  
\end{figure*}

\paragraph{Text-Based Image Editing}
Text-based image editing allows modifications via natural language prompts. Early influential works leveraged the latent space of Generative Adversarial Networks (GANs), such as StyleCLIP~\cite{styleclip}, to perform impressive edits guided by text. With the advent of diffusion models, the field has seen a surge in methods offering higher fidelity and flexibility. For instance, DiffusionCLIP~\cite{diffclip} utilizes CLIP to fine-tune diffusion models for high-quality zero-shot edits across domains. Imagic~\cite{imageic} optimizes text embeddings and diffusion models for complex semantic changes, preserving image detail. InstructPix2Pix~\cite{instructp2p} integrates a language model with a text-to-image model for fast instruction-based edits. Null-text Inversion~\cite{nulltext} refines edits by optimizing null-text embeddings, avoiding weight adjustments. GLIDE~\cite{glide} employs text-conditional diffusion for realistic editing. InfEdit~\cite{infedit} enhances control over intricate edits like texture using inverse techniques. Text-based methods excel in semantic flexibility but lack precise spatial control.

\paragraph{Drag-Based Image Editing}
Drag-based image editing enables precise control through user drag inputs. DragGAN~\cite{dragan} employs GANs with point tracking to adjust poses and shapes, though its quality trails diffusion models. DragDiffusion~\cite{dragdiff} adapts diffusion for point-based edits via latent optimization. FastDrag~\cite{fastdrag} speeds up drag edits, balancing efficiency and quality for real-time use. StableDrag~\cite{stabledrag} enhances diffusion-based motion edits for stability. DragonDiffusion~\cite{dragondiff} uses feature correspondences for object adjustments without fine-tuning. LucidDrag~\cite{lucidrag} and DragText~\cite{dragtext} introduce text control to refine dragging, yet primarily serve drag precision rather than simultaneous semantic and pose editing. These methods achieve fine posture and layout tweaks but often sacrifice full text-driven semantic control, limiting dual editing capabilities. 
\section{Point-Cloud Deterministic Drag}\label{Point-Cloud Deterministic Drag}
\paragraph{Overview.} The architecture of TDEdit is illustrated in \Fig{TDEdit}. Following the paradigm of latent diffusion, we first encode the input image into a latent representation, while estimating its depth using Depth Anything v2~\cite{da2} to guide the subsequent PCDD. The PCDD module then warps the latent features using the rescaled depth and drag conditions to align with user intent. Finally, our DTGD denoiser refines the distorted latent using both original features and source/target text prompts for high-fidelity generation.

The DTGD architecture follows InfEdit and includes three branches for precise control: (1) a source branch, which uses the source prompt and initial latent to steer the denoising process; (2) a reference branch, guided by the target prompt and initial latent, to generate semantic details while preserving the source layout; and (3) a target branch, which integrates drag information and details from the other branches using the target prompt and the warped latent from PCDD. DTGD leverages Hybrid Attention Control to manage inter-branch interactions during noise prediction, combining Blended Layout Control for global layout alignment and local drag retention, and Reference Detail Injection for maintaining semantic and detail fidelity. This process ultimately produces a denoised image that aligns with both the drag instructions and the target text description.  

Before delving into the details of the denoising process, we first explain our core contribution for geometric control: the Point-Cloud Deterministic Drag (PCDD) module. Assume $(x_s,y_s)$ is a position in the original feature $Z_{T}\in\mathbb{R}^{h\times w\times d}$; PCDD seeks to estimate its corresponding position after dragging, $({x_t},{y_t})$, to warp the latent feature accordingly.

\subsection{3D Point Cloud Construction}\label{Build 3D Coordinates}
To model complex deformations realistically, our first step is to transform the 2D problem into a 3D space. 

\paragraph{Depth Map Normalization.} For the original image $I$, we use the monocular depth estimation model Depth Anything V2 to obtain its corresponding depth map $DP_I$. As our operations occur in the latent space, this depth map is resized to match the latent dimensions $(h,w)$ and its values are normalized to a consistent scale, specifically to the range $[\text{dp}_{\min}, \text{dp}_{\max}]$ (default $[0, 63]$), using the following formula:
\begin{equation}\label{supp:depth_rescalation}
   DP'_{Z_T} = \text{dp}_{\min} + \frac{\mathrm{Resize}(DP_I, (h,w)) - DP_{\min}}{DP_{\max} - DP_{\min}} \cdot (\text{dp}_{\max} - \text{dp}_{\min}).
\end{equation}
This allows us to lift every 2D point $(\hat{x}, \hat{y})$ within the user-provided $Mask$ into a 3D point cloud $\hat{P}$ by assigning its depth value as the z-coordinate.

\paragraph{Local Coordinate System.} To simplify subsequent calculations, we establish a local coordinate system centered at the object's origin $\hat{O}$. This origin is estimated from the mask's centroid and then calibrated with a slack distance $d_O$ to better approximate the true center of rotation. All points $p$ are converted to local coordinates via translation: $p_{local} = p_{global} - \hat{O}$.  

\paragraph{Drag Subject Filtering.} A critical step for precise editing is to isolate the target object. We achieve this by filtering the point cloud based on depth, assuming that points belonging to the same object have similar depth values. We define the set of movable points, $P_{drag}^s$, as those within a depth threshold $d_{\mathrm{shield}}$ of the primary handle point $a_1$. This is formally defined as:
\begin{equation}\label{supp:filter_method}
P_{drag}^s = \{p_i \in P \mid |z_{p_i}-z_{a_1}| \leq d_{\mathrm{shield}}\}.
\end{equation}
All other points within the mask, whose depth suggests they belong to a different object or the background, are considered static, $P_{static}$, and are excluded from transformations. This ensures the drag operation is precisely constrained to the intended region.

\subsection{Hierarchical Drag Intention Modeling}\label{Hierarchical Drag Intention Modeling}
Real-world manipulations often involve a combination of rigid body motion and local shape deformation. A key challenge when interpreting multiple drag instructions, e.g., $(a_1, b_1), (a_2, b_2), \dots$, is the inherent ambiguity in user intent. A naive approach, such as averaging the transformations implied by each pair, can lead to unpredictable or undesirable results, especially when instructions are conflicting.

To resolve this ambiguity and create an intuitive, predictable system, we propose a hierarchical interpretation of user intent. Our core design principle is to treat the first drag pair $(a_1, b_1)$ as the representation of the user's primary, global intention, which dictates the rigid motion of the entire object. Subsequent drag pairs are then interpreted as secondary, local instructions for non-rigid deformation, refining the object's shape after its primary motion is complete. This hierarchical model offers several key advantages. It provides a predictable outcome by establishing a stable and unambiguous basis for the object's global motion. The model also aligns with natural cognitive workflows, mirroring the coarse-to-fine process of human interaction. Finally, its design promotes algorithmic stability by elegantly decomposing the complex task into two more manageable sub-problems: a well-defined rigid body motion determined by a single vector, and a flexible non-rigid deformation field influenced by all control points. Following this principle, we implement a two-stage, coarse-to-fine hybrid drag mechanism that first establishes the object's global position based on $(a_1, b_1)$ and then refines its local shape using all available instructions.

\subsection{Hybrid-Rigid Drag}\label{Hybrid-Rigid Drag}
Following the hierarchical principle outlined above, our hybrid drag mechanism is implemented in two distinct stages: a coarse rigid transformation followed by a fine-grained non-rigid deformation.

\paragraph{Rigid Transformation.} The first stage is a rigid transformation, which achieves the overall movement of the point cloud while maintaining geometric consistency. We use the primary user instruction $(a_1,b_1)$ to define this transformation, which consists of a rotation followed by a translation. The rotation matrix $R$ is constructed using Rodrigues' formula from the drag vector. The final position after this rigid step, $P_{rigid}$, serves as the basis for subsequent refinement and is obtained as:
\begin{equation}
   P_{rigid} = R \cdot P_{drag}^s + \alpha \cdot (b_1 - R \cdot a_1).
\end{equation}
Here, $(b_1 - R \cdot a_1)$ represents the translation vector $V_{tra}$, and $\alpha$ is a scaling factor that controls the effectiveness of the translation.

\paragraph{Non-Rigid Deformation.} The second stage refines the object's shape based on all user instructions. We use Radial Basis Function (RBF) interpolation to model a smooth deformation field. First, we define a set of control points $C=A\cup F$, which includes both the user's handle points $A$ (the "drivers" of the deformation) and a set of automatically determined internal fixed points $F$ that act as structural "anchors" to preserve object integrity. The displacement vector $s(p)$ for any point $p$ is then a weighted sum of influences from all control points, using a multiquadric kernel $\phi$:
\begin{equation}\label{rbf_interpolation}
s( p)=(\sum_{c_k\in C} w_k^x \cdot \phi_{p,c_k},\sum_{c_k\in C} w_k^y \cdot \phi_{p,c_k},\sum_{c_k\in C} w_k^z \cdot \phi_{p,c_k}),
\end{equation}
where the kernel function is defined as:
\begin{equation}\label{multiquadric}
    \phi_{p,q}=\sqrt{1+(\mu||p-q||)^2}.
\end{equation}
Here, $\mu$ is a shape parameter that controls the influence radius of the kernel. The unknown weights $w_k=(w_k^x,w_k^y,w_k^z)$ are determined by solving a system of linear equations derived from the known displacements at the control points. This step is crucial as it translates the user's explicit instructions into a solvable mathematical system. Specifically, handle points must move towards their targets, while fixed points must remain stationary:
\begin{equation}\label{fitting_process}
    s_{c_i}=\begin{cases}b_i - a_i ,&c_i\in A,\\
                        \mathbf{0},&c_i\in F.\\\end{cases}
\end{equation}
To ensure the deformation is localized and does not unrealistically affect distant parts of the object, we introduce a distance-based weight $\gamma(p)$. This function modulates the RBF field's influence, making it strongest near handle points and weakest near fixed points, ensuring a natural falloff effect:
\begin{equation}
    \gamma (p)=\frac{\min_{f_i\in F}\|p-f_i\|}{\min_{a_i\in A}\|p-a_i\|+\min_{f_i\in F}\|p-f_i\|}.
\end{equation}
The final coordinates $P^t_{drag}$ are obtained by additively blending the rigid motion with this localized non-rigid refinement. This combination is controlled by a non-rigid influence weight $\beta$:
\begin{equation}
P^t_{drag} =P_{rigid}+\beta \cdot \gamma(P_{rigid}) \cdot s(P_{rigid}),
\end{equation}
where $V_{non-rigid} = \gamma(P_{rigid}) \cdot s(P_{rigid})$.

\subsection{Feature Mapping and Interpolation}\label{Feature Mapping}
With the final 3D coordinates $P^t_{drag}$ computed, we must apply this transformation back to the 2D latent feature map $\widetilde{Z}_T$. 

\paragraph{3D-to-2D Projection.} First, the continuous 3D coordinates are discretized by rounding the x and y components to align with the latent grid:
\begin{equation}
    \hat{P}^t_{drag}=\{(\text{round}(x_t),\text{round}(y_t),z_t)|(x_t,y_t,z_t)\in P^t_{drag}\}.
\end{equation}
A key challenge in this 3D-to-2D projection is handling self-occlusion. To resolve this, we employ a "z-buffering" strategy: if multiple 3D points map to the same 2D location, only the one with the maximum z-value (i.e., the one in the foreground) is retained. This process results in a sparse set of valid target points. Let $\text{Inv}: (x_t, y_t) \mapsto (x_s, y_s)$ denote the inverse mapping that retrieves the source 2D coordinates $(x_s, y_s)$ for a given target 2D coordinate $(x_t, y_t)$ within the dragged region. The final warped latent map $\widetilde{Z}_T$ is then constructed by combining three types of features: (1) features of dragged points, which are transferred from their corresponding source locations in $Z_T$; (2) features of static points within the mask, which are copied directly; and (3) features of points outside the mask, which also remain unchanged. This comprehensive logic is formally expressed as:
\begin{equation}
\widetilde{Z}_T(x,y) = 
\begin{cases} 
    {Z}_T(\text{Inv}(x,y)), & \text{if }(x,y) \text{ is a dragged point}, \\
    Z_T(x,y),  & \text{otherwise}.
\end{cases}
\end{equation}

\paragraph{Interpolation of Featureless Points.} The direct feature transfer and z-buffering in the previous step inevitably creates voids or "holes" in the warped latent map. To ensure a continuous and semantically coherent latent representation for the subsequent denoising stage, we fill these featureless regions using Bidirectional Nearest Neighbor Interpolation (BNNI), as proposed by FastDrag. For any empty-feature point $(x_{\mathcal{N}},y_{\mathcal{N}})$, its feature is interpolated as a weighted average of its four nearest neighbors (up, down, left, right):
\begin{equation}
\widetilde{Z}_T{\left(x_{\mathcal{N}},y_{\mathcal{N}}\right)}=\sum_{loc=u,r,d,l}w_{loc}\cdot \widetilde{Z}_T(x_{loc},y_{loc}),
\end{equation} 
where the weights $w_{loc}$ are inversely proportional to the distance to each neighbor, giving closer points more influence:
\begin{equation}
w_{loc}=\frac{1/len_{loc}}{\sum_{loc=u,r,d,l}1/len_{loc}}.
\end{equation}
\section{Drag-Text Guided Denoising}\label{Drag and Text Guided Denoising}

The PCDD module injects drag guidance into the warped latent feature $\widetilde{Z}_T$. Our DTGD module then integrates this distorted representation with text conditioning through attention map manipulation in the denoising UNet (as shown in \Fig{TDEdit}), adaptively fusing both conditions to progressively denoise the latent space and generate the final output image.

\subsection{Hybrid Attention Control}\label{Hybrid Attention Control}
\paragraph{Blended Layout Control.} In our practice, a key challenge in combining drag-based and text-based methods lies in layout control. We take three branches and exchange information among three branches via attention maps to solve the problem. Formally, $M^{src},M^{ref},M^{tgt}$ are the attention map from the denoising UNet in source, reference and target branches, respectively, where the image always serves the key and value, and the source prompt (source branch) and target prompt (reference and target branches) act as the query. 

Subsequently, we endow the reference branch with a similar layout to the source one via an attention replacement. Formally, when the $j$-th token in the target prompt matches the $i$-th token in the source prompt, we replace the $j$-th row of $M^{ref}$ with the corresponding $i$-th row from $M^{src}$. The process is as:
\begin{equation}
\mathrm{Replace}(M^{src}, M^{ref})_{j} = 
\begin{cases} 
M^{src}_i & \text{if }\mathcal{A}(j)=i, \\ 
M^{ref}_j & \text{if } \mathcal{A}(j)=\text{None}.
\end{cases}
\end{equation}
where $\mathcal{A}(j)=i$ means the $j$-th token in target prompt matches the $i$-th token in source prompt. To form the rough layout in early timesteps and detailed semantic layout in late timesteps, we set a timestep threshold $t_c$ to control the degree of replacement in the cross-attention process of noise prediction:
\begin{equation}
M^{ref}= \begin{cases} 
\mathrm{Replace}(M^{src}, M^{ref}) & \text{if } t \geq t_c, \\
M^{ref} & \text{if } t < t_c.
\end{cases}
\end{equation}
This modified attention map, $M^{ref}$, is then utilized within the UNet's attention layers during the noise prediction step for the reference branch, contributing to the generation of $Z_t^{ref}$. Subsequently, to ensure the target branch maintains a similar layout in early timesteps without compromising the dragged region, we strategically fuse features:
\begin{equation}
    Z_t^{tgt}=Z_t^{tgt}\odot Mask + Z_t^{ref}\odot (1-Mask).
\end{equation}

\paragraph{Reference Detail Injection.} To introduce detail from the source branch, when performing cross-attention, we replace $(Q^{ref},K^{ref})$ by $(Q^{src},K^{src})$ between reference and source branches in the early timesteps until $t_s$ to prevent excessive detail, and we use $(K^{ref},V^{ref})$ to replace $(K^{tgt},V^{tgt})$ to provide reference throughout the whole process. The detailed formulation is:
\begin{equation}
    \begin{gathered}
        (Q^{ref},K^{ref},V^{ref}) \leftarrow
        \begin{cases}
            (Q^{src},K^{src},V^{ref}) & \text{if } t \geq t_s, \\
            (Q^{ref},K^{ref},V^{ref}) & \text{if } t < t_s,
        \end{cases} \\
        (Q^{tgt},K^{tgt},V^{tgt}) \leftarrow (Q^{tgt},K^{ref},V^{ref}).
    \end{gathered}
\end{equation}
Subsequently, let $l^{src}$ and $l^{tgt}$ be the prompt feature encoded from the language encoder. We can get the noise of three branches from the noise prediction network:
\begin{align}
    \varepsilon_{src} &= \varepsilon_\theta(Z^{src}_t, t, l^{src}),    \label{eq:noise_src} \\
    \varepsilon_{ref} &= \varepsilon_\theta(Z^{ref}_t, t, l^{tgt}),    \label{eq:noise_ref} \\
    \varepsilon_{tgt} &= \varepsilon_\theta(Z^{tgt}_t, t, l^{tgt}).     \label{eq:noise_tgt}
\end{align}
In the following, we use Target-focused DDCM sampling to denoise and get the final target image.

\subsection{Target-focused DDCM Sampling}\label{modified ddcm sampling}
The DDCM sampling formula is a modified DDIM approach for inversion-free image editing, ensuring path consistency between source ($Z^{src}_t$) and target ($Z^{tgt}_t$) branches during denoising:
\begin{equation}\label{modified_ddcm_base}
\begin{aligned}
Z^{tgt}_{t-1} &= \sqrt{\alpha_{t-1}} \left( \frac{Z^{tgt}_t - \sqrt{1 - \alpha_t} \varepsilon}{\sqrt{\alpha_t}} \right) & \text{(predicted } Z_0\text{)} \\
&\quad + \sqrt{1 - \alpha_{t-1} - \sigma_t^2} \cdot \varepsilon & \text{(direction to } Z^{tgt}_t\text{)} \\
&\quad + \sigma_t \varepsilon_t, \quad \text{where } \varepsilon_t \sim \mathcal{N}(0, \boldsymbol{I}) & \text{(random noise)},
\end{aligned}
\end{equation}
where the second term $\sqrt{1 - \alpha_{t-1} - \sigma_t^2} \cdot \varepsilon$ represents the direction toward $Z^{tgt}_t$. The combined noise $\varepsilon=(\varepsilon_{tgt}-\varepsilon_{src}+\varepsilon_{cons})$ reflects the source branch's influence on the target branch. The consistency noise $\varepsilon_{cons}$ is derived from the source branch's latent $Z^{src}_t$ and the initial input $Z_0$, enforcing structural alignment, and is formally defined as:
\begin{equation}
    \varepsilon_{cons} = \frac{Z^{src}_t - \sqrt{\alpha_t} Z_0}{\sqrt{1-\alpha_t}}.
\end{equation}
The term $\varepsilon$ ensures the target branch adapts to editing while staying consistent with the source.

The standard DDCM eliminates the second term by setting $\sigma_t = \sqrt{1 - \alpha_{t-1}}$. However, this strategy weakens the dependence on $Z^{tgt}_t$, causing imprecise layout control from the drag condition. To remedy this issue, we introduce a factor $\eta$ to adjust the dependency on $Z^{tgt}_t$, setting $\sigma_t = \eta \sqrt{1 - \alpha_{t-1}}$. Consequently, our denoising formula reads:
\begin{equation}
\begin{aligned}
Z^{tgt}_{t-1} &= \sqrt{\alpha_{t-1}} \left( \frac{Z^{tgt}_t - \sqrt{1 - \alpha_t} \varepsilon}{\sqrt{\alpha_t}} \right) & \text{(predicted } Z_0\text{)} \\
&\quad + \sqrt{(1-\eta^2)(1 - \alpha_{t-1})} \cdot \varepsilon & \text{(direction to } Z^{tgt}_t\text{)} \\
&\quad + \sigma_t \varepsilon_t, \quad \text{where } \varepsilon_t \sim \mathcal{N}(0, \boldsymbol{I}) & \text{(random noise)}.
\end{aligned}
\end{equation}

To reserve the dragged layout, prevent it from being treated as noise and overly corrected, we set the noise control parameter $\eta<1$ to enhance dependence on $Z^{tgt}_t$, and reduce random noise to further optimize the denoising effect. We implement a dynamic adjustment strategy for $\eta$.

\begin{figure*}[t!] 
\centering 
\includegraphics[width=\textwidth]{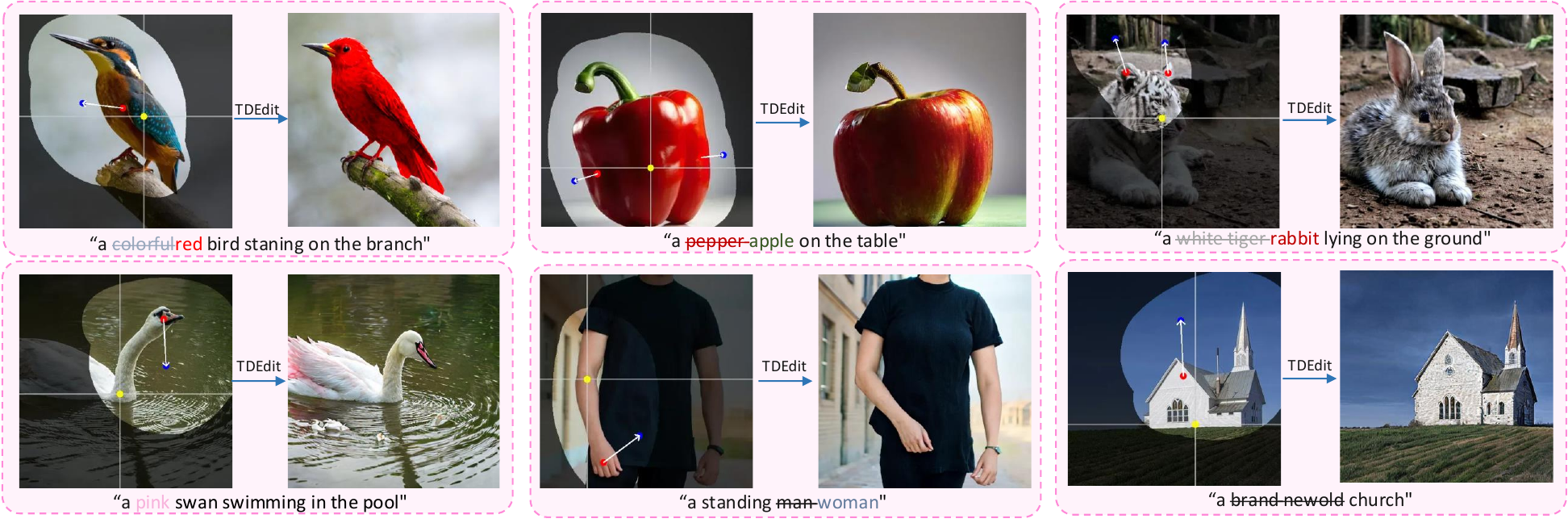}
\caption{Results of Text-Drag joint editing with our TDEdit. The final images clearly follow both conditions and are authentic enough, revealing TDEdit well balances both control signals and synthesize plausible results.} 
\label{Text-Drag} 
\end{figure*}

\paragraph{Dynamically adjust $\eta$.} We adjust $\eta$ based on the progress of the current timestep $t$, the strategy is as:
\begin{equation}
    \eta = \begin{cases} 0.5 &  \text{if } t/T < 0.3 ,
    \\ 0.5 + 0.4 \cdot \frac{ t/T - 0.3}{0.4} & \text{if } 0.3 \leq  t/T \leq 0.7 ,
    \\ 0.9 &   \text{if } t/T > 0.7,
    \end{cases}
\end{equation}
where $t/T$ represents the progress in the total denoising steps. This dynamic adjustment strategy is divided into three stages. In the early stage ($t/T < 0.3$), $\eta$ is set to 0.5 to preserve more drag-induced details. In the middle stage ($0.3 \le t/T \le 0.7$), $\eta$ increases linearly from 0.5 to 0.9, gradually introducing more random noise to smooth the denoising process. In the final stage ($t/T > 0.7$), $\eta$ is held at 0.9 to maintain a sufficient level of randomness, which is crucial for ensuring the smoothness and detail consistency of the final generated image. These specific thresholds and values were determined empirically to achieve a robust balance between geometric fidelity and semantic quality across a wide range of editing tasks.

Both the target and reference branches are progressively denoised using this Target-focused DDCM sampling process. Ultimately, the fully denoised target latent, $Z^{tgt}_0$, is decoded by the VAE to yield the final, high-fidelity edited image.

\section{Experiments}\label{Experiments}

\subsection{Implementation Details}
\paragraph{Experimental Setup.} Our framework is built upon a pretrained Latent Consistency Model (LCM), specifically the LCM Dreamshaper v7 checkpoint~\cite{LCM}. All experiments, including timing benchmarks, were conducted on a single NVIDIA RTX 4090 GPU. For the diffusion process, we perform editing over 10 effective denoising steps, derived from 15 total inference steps with an inversion strength of 0.7. All images are processed at a resolution of 512$\times$512 pixels.

\paragraph{Denoising Process Hyperparameters.} We utilize Classifier-Free Guidance (CFG)~\cite{CFG}, setting the scale to 1.0 for the source branch and 2.0 for the target branch to balance fidelity and edit strength. The reference branch guidance is adaptively set to 1.0 or 2.0 depending on the edit complexity. For our attention control mechanisms, Blended Layout Control is active for the first 3 denoising steps, while for Reference Detail Injection, the initial 5 steps prioritize source feature integration. In our Target-focused DDCM Sampling, the noise scaling parameter $\eta$ is fixed to 1.0 for the source and reference branches, while the target branch's $\eta$ is dynamically adjusted from 0.5 to 0.9 as detailed in \Sec{modified ddcm sampling}. To protect the background, we fuse target features (inside the mask) with reference features (outside the mask) for the first 4 denoising steps.

\paragraph{PCDD Module Hyperparameters.} The origin for rotation and deformation is set as the mask's centroid. The default slack distance $d_O$ used to calibrate the origin is set to 20, and the shield distance $d_\text{shield}$ for isolating the drag subject is 30. The influence weights for rigid transformation ($\alpha$) and non-rigid deformation ($\beta$) are both set to 0.7.

\begin{figure*}[t] 
\centering 
\includegraphics[width=0.95\textwidth]{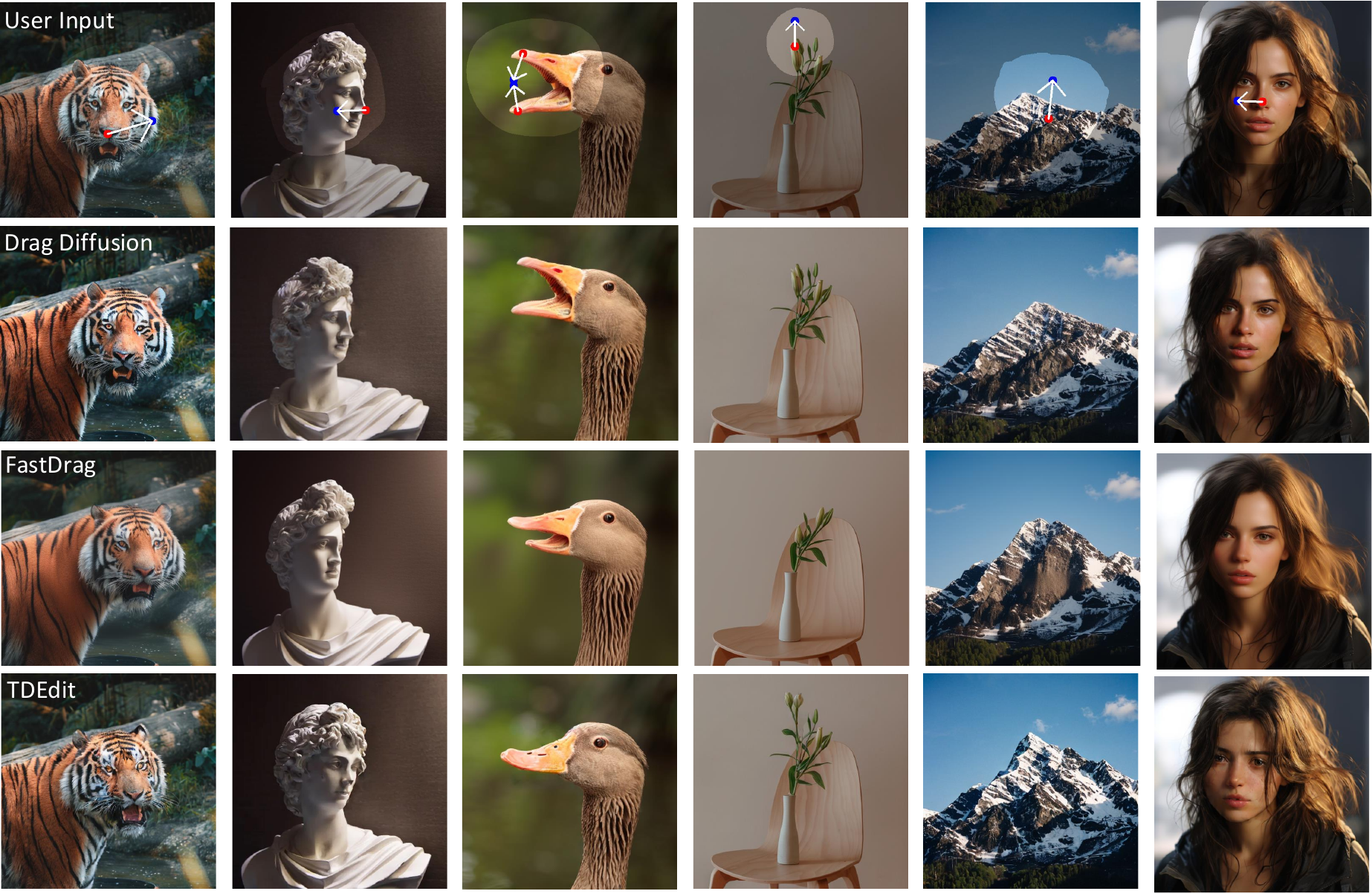}
\caption{Qualitative comparison regarding drag control. In comparison with other models, TDEdit achieves more precise layout drag, and the results of TDEdit well follows the drag intention.} 
\label{Drag} 
\end{figure*}

\subsection{User Study}
\label{sec:user_study}
The novel task of text-drag joint editing presents a unique evaluation challenge, as no established benchmarks currently exist to quantitatively measure the synergistic performance of both modalities. Therefore, to directly assess our framework's effectiveness and user-perceived quality, we conducted a comprehensive user study. The study involved 23 participants recruited from university students and researchers familiar with generative AI tools. It was systematically divided into three parts to evaluate TDEdit's performance in its core scenario of text-drag joint editing, and to benchmark its competitiveness against specialized methods in drag-only and text-only editing scenarios.

\subsubsection{Text-Drag Joint Editing Evaluation}
In this section, we evaluated TDEdit's ability to handle and harmonize simultaneous text and drag instructions. Participants were shown 25 joint editing results generated by TDEdit and were asked to rate each output across four key dimensions on a 5-point Likert scale (1 = Very Poor, 5 = Very Good). The mean scores, standard deviations, and 95\% confidence intervals (CI) are summarized in \Table{tab:joint_study}.

\paragraph{Analysis:}
The results in \Table{tab:joint_study} demonstrate that TDEdit received consistently high scores across all evaluation dimensions, with all mean scores exceeding 4.3. Notably, it achieved a high score of \textbf{4.42} on "Joint Adherence," the most critical metric for our unified framework. The tight 95\% confidence interval of [4.31, 4.53] for this metric strongly indicates that our framework reliably performs synergistic editing. Furthermore, the high "Image Quality" score of 4.34 confirms that TDEdit produces visually plausible and high-fidelity results.

\subsubsection{Drag-Only Editing Comparison}
We conducted a head-to-head comparison against two leading specialized methods: FastDrag~\cite{fastdrag} and DragDiffusion~\cite{dragdiff}. Participants were asked a forced-choice question: "Which result best follows the drag instruction while maintaining the highest image quality?". The results are aggregated in \Table{tab:drag_study}, where Preference Rate (Pref.) indicates the percentage of times a method was chosen as the best.

\paragraph{Analysis:}
As shown in \Table{tab:drag_study}, TDEdit emerged as the most preferred method (40.6\%). While its lead over FastDrag (38.3\%) is narrow, it significantly surpasses the classic DragDiffusion (21.1\%), with their 95\% CIs being clearly disjoint. This provides strong evidence that TDEdit is highly competitive in its individual modalities.

\subsubsection{Text-Only Editing Comparison}
Finally, to evaluate how our modifications for drag-editing impact text-only performance, we conducted a crucial comparison between TDEdit and its architectural foundation, InfEdit~\cite{infedit}. As InfEdit is a powerful, specialized method for text-driven editing, this head-to-head evaluation assesses whether our unified framework successfully retains (or even enhances) the strong text-editing capabilities of the base model. Participants were asked to choose which of the two results better reflected the textual edit while preserving higher overall quality. The user preference statistics are presented in \Table{tab:text_study}.

\begin{table}[t]
    \centering 
    \caption{User study scores for TDEdit on joint editing tasks. Scores are averaged over 23 users for 25 editing cases. Higher is better (out of 5).}
    \label{tab:joint_study}
    \setlength{\tabcolsep}{4pt}
    \begin{tabular}{@{} l S[table-format=1.2] S[table-format=1.2] c @{}}
        \toprule
        \textbf{Dimension} & {\textbf{Score}} & {\textbf{Std. Dev.}} & \textbf{95\% CI} \\ 
        \midrule
        Text Adherence & 4.51 & 0.81 & [4.41, 4.61] \\
        Drag Adherence & 4.49 & 0.86 & [4.38, 4.60] \\
        Joint Adherence & 4.42 & 0.90 & [4.31, 4.53] \\
        Image Quality & 4.34 & 0.92 & [4.23, 4.45] \\ 
        \bottomrule
    \end{tabular}
\end{table}

\begin{table}[t]
    \centering 
    \caption{User preference rates (\%) in the drag-only editing comparison. Total votes were collected from 23 users over 15 cases (345 total votes).}
    \label{tab:drag_study}
    \setlength{\tabcolsep}{4pt}
    \begin{tabular}{@{} l S[table-format=3.0, detect-weight] S[table-format=2.1, detect-weight] c @{}}
        \toprule
        \textbf{Method} & {\textbf{Total Votes}} & {\textbf{Pref.}} & \textbf{95\% CI} \\ 
        \midrule
        \textbf{TDEdit (Ours)} & \bfseries 140 & \bfseries 40.6 & \textbf{[35.4\%, 45.8\%]} \\
        FastDrag & 132 & 38.3 & [33.1\%, 43.5\%] \\
        DragDiffusion & 73 & 21.1 & [21.1\%, 25.6\%] \\
        \bottomrule
    \end{tabular}
\end{table}

\begin{table}[t]
    \centering 
    \caption{User preference rates (\%) in the text-only editing comparison. Total votes were collected from 23 users over 10 cases (230 total votes).}
    \label{tab:text_study}
    \setlength{\tabcolsep}{4pt}
    \begin{tabular}{@{} l S[table-format=3.0, detect-weight] S[table-format=2.1, detect-weight] c @{}}
        \toprule
        \textbf{Method} & {\textbf{Total Votes}} & {\textbf{Pref.}} & \textbf{95\% CI} \\ 
        \midrule
        \textbf{TDEdit (Ours)} & \bfseries 145 & \bfseries 63.0 & \textbf{[56.7\%, 69.3\%]} \\
        InfEdit & 85 & 37.0 & [30.7\%, 43.3\%] \\ 
        \bottomrule
    \end{tabular}
\end{table}

\paragraph{Analysis:} The data in \Table{tab:text_study} clearly indicates that TDEdit was preferred by a significant margin (63.0\%) with non-overlapping 95\% confidence intervals confirming this preference is statistically significant. This result is particularly insightful when considered alongside the quantitative metrics in \Table{quantitative-PIE-Bench}, where InfEdit shows advantages in fidelity metrics like LPIPS and MSE. We hypothesize that this divergence between user preference and quantitative scores stems from the architectural differences between the two models. TDEdit's dynamic multi-branch attention mechanism appears to be more responsive to Classifier-Free Guidance (CFG), effectively amplifying the strength of the textual edit to produce more visually distinct and impactful results. Conversely, InfEdit's layout branch, which is designed to preserve source features for consistency, may inadvertently constrain the full extent of the edit, leading to results that are higher in pixel-level fidelity but sometimes less semantically aligned with the target prompt. Therefore, the strong user preference suggests that while TDEdit and InfEdit exhibit a trade-off between edit strength and detail preservation, the performance gap in text-editing is not perceived as significant by users. In fact, users tend to favor the more pronounced and semantically clear results generated by our framework. This validates that our unified model remains highly competitive in text-only tasks, offering a compelling alternative to its specialized base model.

\subsection{Qualitative Comparison}
\paragraph{Text-Drag Joint Editing.} As an early exploration for text-drag joint control, no proper models can serve as the comparison methods in this setting. \Fig{Text-Drag} shows the results of our TDEdit with both conditions. We can observe that the results can well follow both control signals, achieving the goals of this paper.

\begin{figure}[t] 
\centering 
\includegraphics[width=0.45\textwidth]{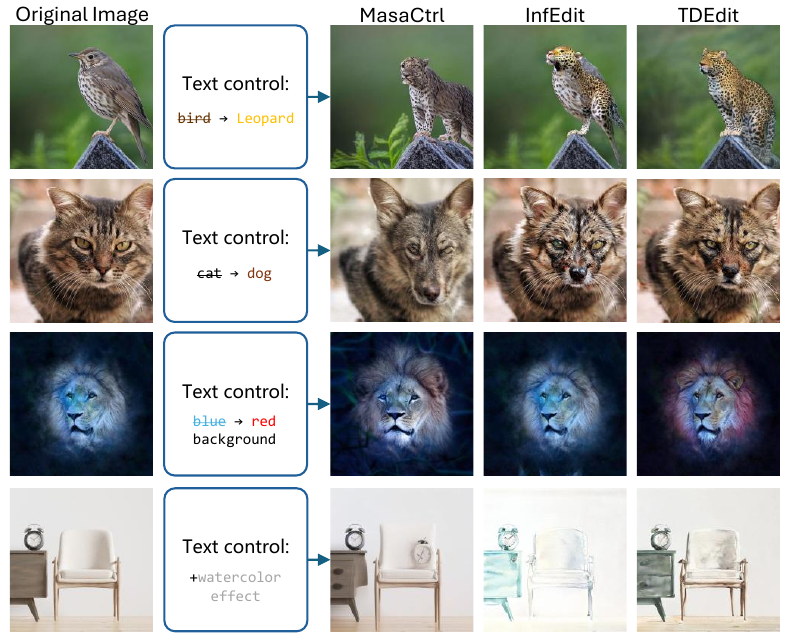}
\caption{Qualitative comparison with InfEdit.} 
\label{Text} 
\end{figure}

\paragraph{Text-Based Editing.}
In the text task, TDEdit outperforms InfEdit in certain aspects. Although TDEdit sacrifices some source branch details for dragging, the quality difference is minor in most cases. As shown in \Fig{Text}, InfEdit's layout branch may retain excessive source features during tasks like replacement, color, and style, hindering alignment with target semantics. TDEdit achieves better generation by incorporating more target semantic information.

\paragraph{Drag-Based Editing.}
In drag tasks, TDEdit surpasses DragDiffusion and FastDrag. DragDiffusion’s n-step optimization causes uncertainty and artifacts, while FastDrag’s one-step mapping lacks realism for precise edits. TDEdit’s 3D feature mapping enables realistic transformations like rotation/stretching (\Fig{Drag}). For details, DragDiffusion and FastDrag lose texture fidelity due to over-smoothing. TDEdit maintains details through attention control and text injection, ensuring semantic consistency.

\begin{table*}[t]
    \centering 
    \small
    \caption{
        Quantitative comparison on the PIE-Bench dataset. For a fair comparison, performance of all baselines is sourced from the InfEdit paper~\cite{infedit}.   
        Notably, our TDEdit, InfEdit, and MasaCtrl employ distinct editing mechanisms, while the other baselines are all based on the Prompt-to-Prompt (P2P)~\cite{prompt2prompt} framework for attention editing.
        For brevity, values for Distance, LPIIPS, MSE, and SSIM are scaled by $10^3$, $10^3$, $10^4$, and $10^2$ respectively.
    }
    \label{quantitative-PIE-Bench}
    \setlength{\tabcolsep}{4pt} 
    \begin{tabular}{@{} l *{7}{S[table-format=3.2, detect-weight]} @{}}
        \toprule
        \textbf{Approach} & 
        {\textbf{Distance $\downarrow$}}& 
        {\textbf{PSNR $\uparrow$}}& 
        {\textbf{LPIPS $\downarrow$}}& 
        {\textbf{MSE $\downarrow$}} & 
        {\textbf{SSIM $\uparrow$}} & 
        {\textbf{CLIP-W $\uparrow$}} & 
        {\textbf{CLIP-E $\uparrow$}} \\
        \midrule
         DDIM~\cite{ddim}& 69.43& 17.87& 208.80& 219.88& 71.14& 25.01& \bfseries 22.44\\
         \textcolor{gray}{CycleD}~\cite{cycled}& 6.06& 28.25& 43.96& 25.85& 85.61& 23.68& 20.87\\
         NT~\cite{nulltext}& 13.44& 27.03& 60.67& 35.86& 84.11& 24.75& 21.86\\
         NP~\cite{NP}& 16.17& 26.21& 69.01& 39.73& 83.40& 24.61& 21.87\\
         StyleD~\cite{StyleD}& \bfseries 11.65& 26.05& 66.10& 38.63& 83.42& 24.78& 21.72\\
         DI~\cite{DI}& \bfseries 11.65& 27.22& 54.55& 32.86& 84.76& 25.02& 22.10\\
         MasaCtrl~\cite{masactrl} & 40.45 & 18.82 & 184.90 & 155.20 & 67.29 & 24.00 & 21.49 \\
         InfEdit~\cite{infedit}& 13.78 & \bfseries 28.51 & \bfseries 47.58 & \bfseries 32.09 & \bfseries 85.66 & \bfseries 25.03 & 22.22 \\
         \midrule
         \textbf{TDEdit (Ours)} & 15.09 & 25.66 & 68.77 & 102.91 & 83.58 & 24.81 & 21.90 \\
        \bottomrule
    \end{tabular}
\end{table*}

\subsection{Quantitative Comparison}
\paragraph{Text-Based Editing.}
We evaluate TDEdit on PIE-Bench~\cite{DI} using a range of standard metrics, including Mean Squared Error (MSE), Peak Signal-to-Noise Ratio (PSNR), Structural Similarity Index Measure (SSIM)~\cite{SSIM}, Learned Perceptual Image Patch Similarity (LPIPS)~\cite{lpips}, a perceptual Distance~\cite{Distance}, and CLIP scores for the whole image (CLIP-W) and the edit region (CLIP-E)~\cite{CLIPScore}. The results are presented in \Table{quantitative-PIE-Bench}. TDEdit shows a slightly higher Distance, LPIPS, and MSE compared to some baselines. This decline stems from our emphasis on supporting drag-based editing. Nevertheless, the impact remains minimal, suggesting editing quality is not significantly compromised.

\begin{table}[t]
    \centering 
    \small
    \caption{Quantitative comparison on DragBench.}
    \label{quantitative-DragBench}
    \begin{tabularx}{0.8\columnwidth}{@{} l >{\centering\arraybackslash}X >{\centering\arraybackslash}X >{\centering\arraybackslash}X @{}} 
        \toprule
        \textbf{Approach} & \textbf{MD $\downarrow$} & \textbf{IF $\uparrow$} & \textbf{Time(s)} \\
        \midrule
        DragDiffusion~\cite{dragdiff} & 33.70 & 0.89 & 21.54 \\
        DragNoise~\cite{dragnoise} & 33.41 & 0.63 & 20.41 \\
        FreeDrag~\cite{freedrag} & 35.00 & 0.70 & 52.63 \\
        GoodDrag~\cite{gooddrag} & \textbf{22.96} & 0.86 & 45.83 \\
        DiffEditor~\cite{diffEdit} & 28.46 & 0.89 & 21.68 \\
        FastDrag~\cite{fastdrag} & 32.23 & 0.86 & 2.80\\
        \midrule
        \textbf{TDEdit (Ours)} & 26.91 & \textbf{0.92} & \textbf{2.52} \\
        \bottomrule
    \end{tabularx}
\end{table}

\paragraph{Drag-Based Editing.}
We conduct quantitative experiments on DragBench~\cite{dragdiff}, where MD \cite{dragan} measures drag precision, and IF~\cite{imageic} (1-LPIPS \cite{lpips}) assesses similarity. "Editing Time" is the total end-to-end duration per instruction. As presented in \Table{quantitative-DragBench}, our TDEdit demonstrates a superior balance of precision, quality, and efficiency. It achieves a competitive Mean Distance (MD) of 26.91, significantly outperforming mainstream methods like DragDiffusion (33.70) and the efficiency-focused FastDrag (32.23). While GoodDrag obtains the lowest MD, its runtime is substantially higher. In contrast, our method not only achieves the highest Image Fidelity (IF) score of 0.92 among all baselines but also operates at the fastest speed of just 2.52 seconds. This combination of strong precision, state-of-the-art image quality, and real-time performance establishes TDEdit as a highly effective and practical solution for drag-based editing.

\subsection{Ablation Study}
\begin{figure}[htbp] \centering \includegraphics[width=0.9\columnwidth]{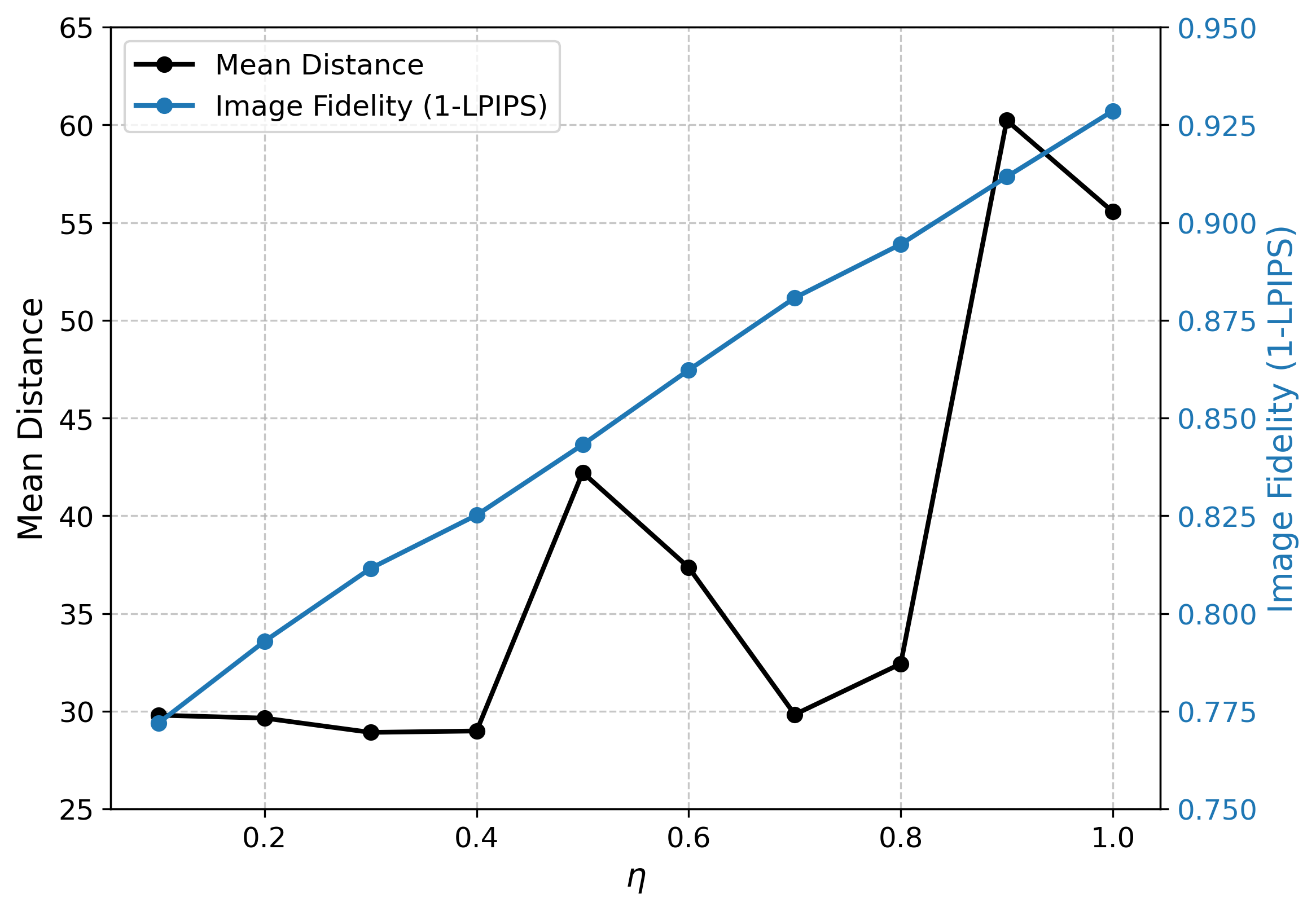}
\caption{Illustration of Ablation Study on $\eta$.} \label{ablation-eta}
\end{figure}

We conduct ablation experiments on Hybrid Attention Control and $\eta$ on DragBench. \Table{ablation-hybrid} shows that Blended Layout Control (BLC) has some impact, while Reference Detail Control has a greater impact on MD and IF, showing the importance of the Reference branch. Our analysis of $\eta$ (\Fig{ablation-eta}) reveals a trade-off between image quality and dragging ease, validating our dynamic adjustment strategy.

\begin{table}[t]
    \centering 
    \small
    \caption{Ablation Study on Hybrid Attention Control.}
    \label{ablation-hybrid}
    \begin{tabular}{
        @{} l 
        S[table-format=2.2, detect-weight] 
        S[table-format=1.2, detect-weight] 
        S[table-format=1.3, detect-weight] 
        @{}
    }
        \toprule
        \textbf{Approach} & {\textbf{MD $\downarrow$}} & {\textbf{IF $\uparrow$}} & {\textbf{CLIP Sim $\uparrow$}} \\
        \midrule
        TDEdit w/o BLC \& RDI & 32.14 & 0.67 & 0.880 \\ 
        TDEdit w/o BLC      & 26.82 & 0.89 & 0.972 \\
        TDEdit w/o RDI      & 31.12 & 0.83 & 0.939 \\
        \textbf{TDEdit (Full)}       & \bfseries 25.89 & \bfseries 0.92 & \bfseries 0.977 \\
        \bottomrule
    \end{tabular}
\end{table}

\section{Limitations}
\label{sec:limitations}

Despite the promising performance of TDEdit as a unified framework, we acknowledge several limitations that offer avenues for future research.

First, the performance of our Point-Cloud Deterministic Drag (PCDD) module can be sensitive to its hyperparameters. Unlike purely 2D drag methods that often only require handle and target points, our 3D-aware approach necessitates adjustments to parameters such as the slack distance ($d_O$) and shield distance ($d_{\text{shield}}$) to achieve optimal results for objects with varying shapes and depths. This introduces a trade-off: while the 3D representation enables more realistic, physically plausible deformations, it currently requires more user expertise to tune for the best effect compared to simpler 2D paradigms.

Second, artifacts can emerge in regions with extensive feature voids. These voids are an inherent consequence of the discrete latent relocation process in drag operations, particularly during large-scale displacements. While our Bidirectional Nearest Neighbor Interpolation (BNNI) strategy effectively fills small gaps by leveraging local neighborhood information, its reconstruction capabilities are limited when confronted with large, contiguous null regions. In such cases, the interpolated features may lack global semantic coherence, leading to visual artifacts or a loss of texture fidelity, as the local nature of BNNI may not fully reconstruct the complex structure of the missing area.

Finally, as our framework builds upon and modifies the architecture of InfEdit~\cite{infedit}, it inherits some of its inherent limitations, such as occasional unintended semantic changes in non-target regions. More importantly, to seamlessly integrate drag-based control, we made a critical adaptation to the original three-branch architecture. Specifically, we repurposed InfEdit’s original layout branch, which was dedicated to fine-grained pose and layout control, into our reference branch, primarily tasked with injecting semantic details from the target text. Concurrently, the core responsibility of our target branch shifted to anchoring and preserving the spatial layout dictated by the drag operation. In InfEdit's original design, the three branches created a delicate balance highly optimized for pure text-driven tasks, where the layout branch was crucial for precisely preserving and adjusting object poses without disturbing non-edited regions. This architectural trade-off, while fundamental to achieving our unified text-drag framework, inevitably disrupts this finely-tuned equilibrium. Specifically, the target latent $Z_t^{tgt}$ is now primarily constrained to faithfully reflect the post-drag structure computed by our PCDD module. This strong constraint can diminish its capacity to freely blend and refine nuanced pose details from the reference latent $Z_t^{ref}$, particularly in non-dragged regions such as the background or secondary objects. This architectural trade-off explains the slight performance drop observed in our quantitative text-editing benchmarks (\Table{quantitative-PIE-Bench}) when compared to the specialized InfEdit model. Addressing these challenges, for instance by developing more sophisticated feature fusion mechanisms or exploring more seamless architectural integrations, remains a key direction for future research.
\section{Conclusion}
This paper introduces a unified diffusion-based framework for image editing that combines text and drag interactions to overcome the limitations of existing methods. While text-driven editing excels in texture manipulation but lacks spatial precision, drag-based editing controls shape/structure but misses texture guidance. The proposed method integrates both. Key innovations include: Point-Cloud Deterministic Drag: Improves spatial control via 3D feature mapping, and Drag-Text Guided Denoising: Dynamically balances text and drag conditions during denoising. The framework supports text-only, drag-only, or combined editing modes while maintaining high fidelity. Experiments show it outperforms or matches specialized single-mode methods, offering a versatile solution for controllable image manipulation.

{
    \small
    \bibliographystyle{ieeenat_fullname}
    \bibliography{sample-base}
}

\end{document}


\title{Supplementary Material for: \\ TDEdit: A Unified Diffusion Framework for Text-Drag Guided Image Manipulation}

\author{Anonymous CVPR submission \\ Paper ID 3532} 

\maketitle
\thispagestyle{empty}

\appendix

\section{Implementation Details}
This appendix provides further details on our experimental setup, the hyperparameters for the diffusion process, and the specific settings for the Point-Cloud Deterministic Drag (PCDD) module.

\subsection{Experimental Setup}
Our framework is built upon the pretrained Latent Consistency Model (LCM), specifically the LCM Dreamshaper v7 checkpoint. The total number of inference steps is set to 15, with a noise strength for inversion of 0.7. This results in 10 effective denoising steps per edit. All experiments were conducted on a single NVIDIA RTX 4090 GPU.

\subsection{Denoising Process Hyperparameters}
\noindent\textbf{Classifier-Free Guidance (CFG).} The source branch's guidance scale is set to 1.0, and the target branch's scale is set to 2.0. The reference branch's guidance scale is adaptive (1.0 or 2.0) based on edit complexity.

\noindent\textbf{Attention Control and Sampling.} Blended Layout Control is active for the first 3 denoising steps. For Reference Detail Injection, the initial 5 steps prioritize source feature integration. In our Target-focused DDCM Sampling, the noise scaling parameter $\eta$ is 1.0 for the source and reference branches, while the target branch's $\eta$ is dynamically adjusted from 0.5 to 0.9. To protect the background, we fuse target features (inside the mask) with reference features (outside the mask) for the first 4 denoising steps.

\subsection{PCDD Module Hyperparameters}
The origin for rotation and deformation is the mask's centroid. The default slack distance $d_O$ is set to 20, and the shield distance $d_\text{shield}$ is 30. Influence weights $\alpha$ (rigid) and $\beta$ (non-rigid) are both set to 0.7.

\section{Detailed PCDD Formulation}
This section provides a detailed mathematical breakdown of the PCDD module to ensure full reproducibility.

\subsection{3D Coordinate System and Point Filtering}
\noindent\textbf{Depth Map Normalization.} Given an input image $I$, we use Depth Anything V2 to get a depth map $DP_I$. We resize it to match the latent dimensions $(h, w)$ and normalize its values to the range $[\text{dp}_{\min}, \text{dp}_{\max}]$ (default $[0, 63]$):
\begin{equation}\label{supp:depth_rescalation}
   DP'_{Z_T} = \text{dp}_{\min} + \frac{\mathrm{Resize}(DP_I, (h,w)) - DP_{\min}}{DP_{\max} - DP_{\min}} \cdot (\text{dp}_{\max} - \text{dp}_{\min}).
\end{equation}

\noindent\textbf{Local Coordinate System.} The origin $\hat{O}$ is estimated from the mask's centroid and calibrated with a slack distance $d_O$. All points $p$ are converted to local coordinates via translation: $p_{local} = p_{global} - \hat{O}$.

\noindent\textbf{Drag Subject Filtering.} To isolate the target object, we define the set of movable points, $P_{drag}^s$, as those whose absolute depth difference from the primary handle point $a_1$ is less than a threshold $d_{\mathrm{shield}}$:
\begin{equation}\label{supp:filter_method}
P_{drag}^s = \{p_i \in P \mid |z_{p_i}-z_{a_1}| \leq d_{\mathrm{shield}}\}.
\end{equation}
Points outside this range are considered static ($P_{static}$).

\subsection{Hybrid-Rigid Drag Formulation}
\noindent\textbf{Rigid Transformation.} The rotation matrix $R$ is constructed using Rodrigues' formula from the primary drag vector $(a_1, b_1)$. The final rigid positions $P_{rigid}$ are:
\begin{equation}
   P_{rigid} = R \cdot P_{drag}^s + \alpha \cdot (b_1 - R \cdot a_1).
\end{equation}

\noindent\textbf{Non-Rigid Deformation.}
The deformation is governed by handle points $A$ and fixed points $F$. We use Radial Basis Function (RBF) interpolation with a multiquadric kernel $\phi_{p,q} = \sqrt{1+(\mu||p-q||)^2}$ to compute a smooth displacement field $s(p)$. The RBF weights are solved from the constraints $s_{a_i} = b_i - a_i$ for handle points and $s_{f_j} = \mathbf{0}$ for fixed points. To localize the deformation, we apply a distance-based weight $\gamma(p)$.

\noindent\textbf{Final Combination.} The final coordinates are:
\begin{equation}
P^t_{drag} = P_{rigid} + \beta \cdot \gamma(P_{rigid}) \cdot s(P_{rigid}).
\end{equation}

\subsection{Feature Mapping and Interpolation}
The final 3D positions $P^t_{drag}$ are projected back to the 2D latent grid, with occlusions resolved via z-buffering. Voids are filled using Bidirectional Nearest Neighbor Interpolation (BNNI).

\section{Detailed DTGD Formulation}
This section provides the detailed mathematical formulations for the Drag-Text Guided Denoising (DTGD) module, as referenced in the main paper.

\subsection{Hybrid Attention Control Details}
\noindent\textbf{Attention Map Replacement.}
In Blended Layout Control, we align the reference branch's layout with the source branch's. Let $M^{src}$ and $M^{ref}$ be the cross-attention maps. If the $j$-th target prompt token matches the $i$-th source prompt token ($\mathcal{A}(j)=i$), we replace the corresponding row in $M^{ref}$:
\begin{equation}\label{supp:attn_replace}
\mathrm{Replace}(M^{src}, M^{ref})_{j} = 
\begin{cases} 
M^{src}_i & \text{if }\mathcal{A}(j)=i, \\ 
M^{ref}_j & \text{otherwise}.
\end{cases}
\end{equation}
This is active for early timesteps ($t \geq t_c$).

\noindent\textbf{Key-Value Replacement.}
In Reference Detail Injection, we control the flow of (Q, K, V) pairs. We replace $(Q^{ref}, K^{ref})$ with $(Q^{src}, K^{src})$ for $t \geq t_s$, and replace $(K^{tgt}, V^{tgt})$ with $(K^{ref}, V^{ref})$ for all timesteps:
\begin{equation}\label{supp:kv_replace}
    \begin{gathered}
        (Q^{ref},K^{ref},V^{ref}) \leftarrow
        \begin{cases}
            (Q^{src},K^{src},V^{ref}) & \text{if } t \geq t_s, \\
            (Q^{ref},K^{ref},V^{ref}) & \text{if } t < t_s,
        \end{cases} \\
        (Q^{tgt},K^{tgt},V^{tgt}) \leftarrow (Q^{tgt},K^{ref},V^{ref}).
    \end{gathered}
\end{equation}
The noise predictions for each branch are then computed as:
\begin{align}
    \varepsilon_{src} &= \varepsilon_\theta(Z^{src}_t, t, l^{src}), \\
    \varepsilon_{ref} &= \varepsilon_\theta(Z^{ref}_t, t, l^{tgt}), \\
    \varepsilon_{tgt} &= \varepsilon_\theta(Z^{tgt}_t, t, l^{tgt}).
\end{align}

\subsection{Target-focused DDCM Sampling Details}
The standard DDCM sampling step is given by:
\begin{equation}
\begin{split}
Z^{tgt}_{t-1} = \sqrt{\alpha_{t-1}} \left( \frac{Z^{tgt}_t - \sqrt{1 - \alpha_t} \varepsilon}{\sqrt{\alpha_t}} \right) \\
+ \sqrt{1 - \alpha_{t-1} - \sigma_t^2} \cdot \varepsilon + \sigma_t \varepsilon_t.
\end{split}
\end{equation}
The combined noise term $\varepsilon$ is defined as $\varepsilon = \varepsilon_{tgt} - \varepsilon_{src} + \varepsilon_{cons}$, where the consistency noise $\varepsilon_{cons}$ is:
\begin{equation}
    \varepsilon_{cons} = \frac{Z^{src}_t - \sqrt{\alpha_t} Z_0}{\sqrt{1-\alpha_t}}.
\end{equation}
Our key modification is setting $\sigma_t = \eta \sqrt{1 - \alpha_{t-1}}$, where $\eta$ is dynamically adjusted according to the following schedule:
\begin{equation}\label{supp:eta_schedule}
    \eta = \begin{cases} 
        0.5 & \text{if } t/T < 0.3, \\
        0.5 + 0.4 \cdot \frac{ t/T - 0.3}{0.4} & \text{if } 0.3 \leq t/T \leq 0.7, \\
        0.9 & \text{if } t/T > 0.7.
    \end{cases}
\end{equation}
This schedule preserves drag geometry early on and refines details later.